\documentclass[10pt,twocolumn,letterpaper]{article}

\usepackage{cvpr}
\usepackage{times}
\usepackage{epsfig}
\usepackage{graphicx}
\usepackage{amsmath}
\usepackage{amssymb}
\usepackage{cuted}
\usepackage{comment}

\usepackage[pagebackref=true,breaklinks=true,letterpaper=true,colorlinks,bookmarks=false]{hyperref}

 \cvprfinalcopy 

\def\cvprPaperID{1461} 

\ifcvprfinal\fi

\usepackage[utf8]{inputenc}

\title{Motion Prediction Under Multimodality with Conditional Stochastic Networks}
\author{
Katerina Fragkiadaki \\
Google,CMU \\
\textit{katef@cs.cmu.edu}
\and Jonathan Huang \\
Google \\
\textit{jonathanhuang@google.com}\\
\and Alex Alemi \\
Google \\
\textit{alemi@google.com}\\
\and Sudheendra Vijayanarasimhan  \\
Google \\
\textit{svnaras@google.com}\\
\and Susanna Ricco \\
\textit{ricco@google.com}\\
Google \\
\and Rahul Sukthankar \\
Google \\
\textit{sakthankar@google.com}\\
}

\author{
Katerina Fragkiadaki \thanks{Carnegie Mellon University} \\
{\tt\small katef@cs.cmu.edu}
\and 
Jonathan Huang \thanks{Google Research} \\
{\tt\small jonathanhuang@google.com}
\and Alex Alemi \footnotemark[2] \\
{\tt\small alemi@google.com}
\and Sudheendra Vijayanarasimhan  \footnotemark[2] \\
{\tt\small svnaras@google.com}
\and Susanna Ricco \footnotemark[2] \\
{\tt\small ricco@google.com} 
\and Rahul Sukthankar \footnotemark[2] \\
{\tt\small sakthankar@google.com}\\
}

\newcommand{\tr}{\mathrm{tr}}
\newcommand{\Hor}{\mathrm{h}}

\newcommand{\ncodes}{d}
\newcommand{\nsamples}{\mathrm{K}}

\newcommand{\xx}{x}
\newcommand{\yout}{y}

\newcommand{\yhat}{\hat{y}}

\newcommand{\trx}{\mathrm{x}}
\newcommand{\try}{\mathrm{y}}
\newcommand{\ea}{\textit{et al.}}

\newcommand{\encoder}{\mathrm{E}}
\newcommand{\decoder}{\mathrm{D}}
\newcommand{\zz}{z}

\newcommand{\MW}{W_\mu}
\newcommand{\SW}{W_\sigma}
\newcommand{\Mb}{b_\mu}
\newcommand{\Sb}{b_\sigma}
\newcommand{\N}{N}
\newcommand{\Nf}{N^f}

\begin{document}

\maketitle



\begin{abstract}
Given a visual history,  
multiple  future  outcomes for a video scene are  equally probable, in other words,  
the distribution of future outcomes has multiple modes.  Multimodality is notoriously hard to handle by standard regressors or classifiers: the former regress to the mean and the latter discretize a continuous high dimensional output space.  
In this work, we present  stochastic neural network architectures 
that handle such multimodality through stochasticity: future  trajectories of objects, body joints or frames are  represented as deep,  non-linear transformations of  random (as opposed to deterministic) variables. 
Such random variables are sampled from simple Gaussian distributions whose means and variances are parametrized by the output of  convolutional encoders over the visual  history. We introduce novel convolutional   architectures for predicting future body  joint trajectories that outperform fully connected alternatives \cite{DBLP:journals/corr/WalkerDGH16}. We 
introduce stochastic spatial transformers through optical flow warping for predicting  future frames, which outperform  their deterministic equivalents  \cite{DBLP:journals/corr/PatrauceanHC15}.  
Training  stochastic networks involves an  intractable marginalization over stochastic variables. We compare various training schemes that handle such marginalization through a) straightforward sampling from the prior, b) conditional variational autoencoders  \cite{NIPS2015_5775,DBLP:journals/corr/WalkerDGH16},  and, c) a proposed K-best-sample loss that penalizes the best prediction  under a fixed  `` prediction budget". We show  experimental results on object trajectory prediction, human body joint trajectory prediction and video prediction under  varying future uncertainty, validating quantitatively and qualitatively our architectural choices and training schemes. 
\end{abstract}

\section{Introduction}
\label{sec:introduction}


Humans live in the future: we constantly predict what we are about to perceive (see, hear or feel)  during our interactions with the world, and learn from the deviations of our expectations from the reality \cite{de-Wit8702}. 
While watching a scene, we are perfectly happy with one of its \textit{many} possible future evolutions, 
but unexpected outcomes cause surprise.  Predictive computational models should  similarly be able to \textit{predict any of the  different modes of the  distribution of future outcomes}.

\begin{figure}[t!]
\begin{center}
\includegraphics[width=0.45\textwidth]{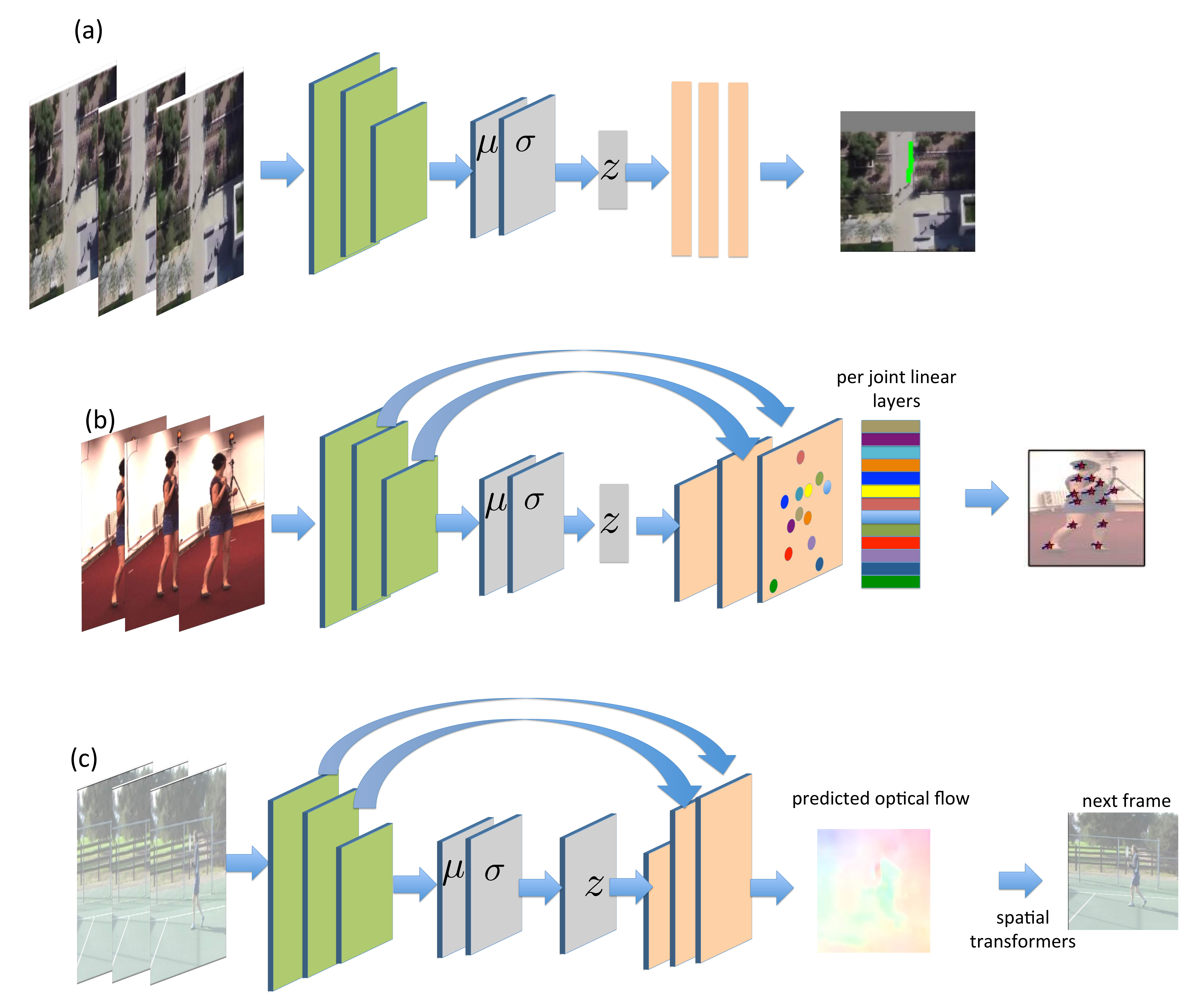}
\end{center}
\caption{\textbf{Conditional stochastic networks} for object trajectory forecasting (a), body joint trajectory forecasting (b) and video forecasting (c). Encoder,  stochastic and decoder subnetworks are depicted in green, gray and orange, respectively. 
\textit{(a):} Visual glimpses are centered around the object of interest. Glimpses here depict overhead camera views. 
\textit{(b):}  \textit{Conv-indexed decoders}  select feature vectors from a deconvolution network based on pixel coordinates, and predict each body joint trajectory using a separate linear layer on top of the selected feature vector. 
\textit{(c):} Samples of future dense motion fields (future optical flows) differentiably warp the current frame to generate samples of the future one.
}
\label{fig:Pnets}
\end{figure}

We  present stochastic neural network architectures that predict samples of  future outcomes conditioned on a history of visual glimpses, and apply them to the tasks of object trajectory, body joint trajectory  and  frame forecasting (Figure \ref{fig:Pnets}).  
Stochastic neural networks are networks where some of the activations are not deterministic but rather sampled from (often Gaussian)  distributions. Such normally distributed random variables are decoded after a series of convolutions and non-linear layers to highly multimodal distributions of the output space. 
In our work,  the means and variances of the Gaussian distributions  are deep non-linear transformations of the network input (the past visual history), that is, the stochastic layer is ``sandwiched" between deterministic encoder and decoder sub-networks, generalizing previous architectures of  \cite{DBLP:journals/corr/WalkerDGH16,NIPS2016_6552} which sampled from Gaussian unit variance noise. Once trained, sampling  the stochastic variables and decoding them  provides plausible samples of the future outcomes.

We present novel architectures and  training schemes for stochastic neural networks.   
For forecasting human body joint trajectories we introduce  \textit{conv-indexed  decoders}, depicted in Figure \ref{fig:Pnets}(b).   
Previous works on body joint trajectory prediction  use fully connected layers since joints are present only on a handful of pixels, not at every pixel \cite{DBLP:journals/corr/WalkerDGH16}. This limits  generalization performance as each trajectory depends on the whole body configuration.  We instead use fully convolutional networks and index into the final convolutional feature map using the body joint pixel coordinates  -that are assumed known for the forecasting task. We thus predict body trajectories with a set of linear layers, one per body joint, each conditioned on a different feature embedding, extracted from the corresponding pixel location of the deconvolutional map.  
We empirically show that the proposed "as-convolutional-as possible" architecture outperforms by a margin fully connected alternatives \cite{DBLP:journals/corr/WalkerDGH16}.  Second, for frame prediction, we present \textit{stochastic dense spatial transformers}:  we compose the future video frame through prediction of a dense pixel motion field (optical flow) and differentiable backward warping that essentially re-arranges pixels of the input frame. Different samples from our model result in different future motion fields and corresponding frame warps. 

Training stochastic networks involves an intractable marginalization over the stochastic variables. We  present and compare  three  training schemes  that handle in different ways such intractable marginalization: a)  Straightforward   sampling for the prior. We simply collect samples from our Gaussian variables for each training example and maximize the empirical likelihood. b) Variational approximations, a.k.a. \textit{conditional variation autoencoders}  that sample from an approximate posterior distribution represented by a recognition model that has access to the output we want to predict. During training we minimize the Kullback–Leibler (KL) divergence between such informed posterior distribution and the prior  distribution provided by our generator model. c) $K$-best-loss, a loss we introduce in this work, which penalizes only \textit{the best  sample}  from a fixed ``prediction budget" for  each training example and training iteration. 

We evaluate the proposed architectures and training schemes on  three forecasting tasks:  (1) predicting motion trajectories (of pedestrians, vehicles, bicyclists etc.) from overhead cameras in the Stanford drone 
dataset~\cite{robicquet2016forecasting}, (2) predicting future body joint motion trajectories
in the H3.6M dataset~\cite{h36m_pami},  and predicting future video frames ``in the wild''  using Freiburg-Berkeley Motion Segmentation~\cite{springerlink:10.1007/978-3-642-15555-0_21} and H3.6M datasets. We evaluate each training scheme under varying future  uncertainty: the longer the frame history we condition upon, the lesser the future uncertainty, the fewer the modes in the output distribution.  We show $K$-best-loss can better handle situations of \textit{high uncertainty}, with many \textit{equiprobable} outcomes, such as when predicting the future from a single image.  
We show extensive quantitative and qualitative results on paper and on our \href{https://sites.google.com/site/conditionalstochasticnets/}{online website}  
which confirm the validity of architectural innovations and facility of proposed training schemes.

The closest works to ours are the work of Walker et al.  \cite{DBLP:journals/corr/WalkerDGH16} that predicts dense future trajectories from a single image and the work of Xue et al. \cite{NIPS2016_6552} that predicts a future frame given a pair of frames.   Both use stochastic units by sampling zero mean and unit variance Gaussian noise concatenated with the input feature maps and train using conditional variational autoencoders. Our work extends those in the following ways: a) proposes indexed convolutional decoders in place of fully connected decoders for trajectory forecasting and shows dramatic improvements, b)  proposes a novel $K$-best for training stochastic networks under extreme uncertainty, c) analyses the training  difficulties by comparing multiple training schemes under varying uncertainty, d) proposes a simpler architecture than \cite{NIPS2016_6552} based on optical flow warping for frame prediction and e) applies stochastic networks in diverse tasks and datasets.

\section{Related Work}

\paragraph{Multimodality in forecasting}
Handling multimodality in prediction is a notoriously difficult problem \cite{Theis2016a}. Models that minimize standard regression losses  perform poorly as they tend to regress towards a mean (blurry) outcome \cite{DBLP:journals/corr/MathieuCL15}.  Models that combine regression and adversarial losses \cite{DBLP:journals/corr/MathieuCL15} suffer  less from regression to the mean, but \textit{latch onto one mode of the distribution and neglect the rest},  as noted in \cite{Theis2016a}. Classifiers that employ  softmax losses need to discretize potentially  high-dimensional output spaces, such as pixel motion fields, and suffer from the corresponding discretization errors. Mixture component networks \cite{DBLP:journals/corr/Graves13,tuske_icassp2015} parametrize a Gaussian Mixture Model (GMM) (mixture components weights, means and variances) where  each mixture represents a different outcome. They have been  used successfully for hand writing generation in \cite{DBLP:journals/corr/Graves13} to predict the next $(x,y)$ pen stroke coordinate, given the generated writing so far. However, it is hard to train many mixtures of high dimensional output spaces, and, as it has been observed, many components often remain un-trained, with one component dominating the rest \cite{Fragkiadaki_2015_ICCV}, unless careful mixture balancing is designed \cite{DBLP:journals/corr/ShazeerMMDLHD17}. 

Many recent data driven approaches predict motion directly from image pixels. 
In \cite{walker2014patch}, 
a large, nonparametric image patch vocabulary is built for patch motion regression. In \cite{DBLP:journals/corr/WalkerGH15}, dense  optical flow is predicted from a single image and the multimodality of motion is handled by considering a different softmax loss for every pixel. 
Work of \cite{fragkiadaki2015learning} predicts ball trajectories in synthetic ``billiard-like'' worlds directly from a sequence of visual glimpses using a regression loss. Work of \cite{Fragkiadaki_2015_ICCV} uses recurrent networks to predict a set of body joint heatmaps at a future frame. Such representation though cannot possibly group the heatmap peaks into coherent 2D pose proposals.  Work of \cite{nal-vpn-2016}  casts frame prediction as sequential conditional prediction, and samples from a categorical distribution of 255 pixel values at every pixel location,  conditioning at the past history and image generated so far. It is unclear  
how to handle  the computational overhead of such models effectively.

\paragraph{Stochastic neural networks.}
Stochastic variables have been used in a variety of settings in the deep learning literature
e.g., for generative modeling, regularization, reinforcement learning, etc.  A key issue is how to train networks with stochastic
variables.  Restricted Boltzmann machines (RBMs), for example~\cite{hinton2006fast} are typically trained using MCMC sampling.  REINFORCE learning rules  were introduced by Willliams~\cite{williams1992simple} in order to  compute sample approximations to gradients in reinforcement learning settings and have also been useful in learning hard attention mechanisms~\cite{mnih2014recurrent,xu2015show,yeung2015end}.
In recent years papers have also proposed stochastic units that can be trained using ordinary SGD, with Dropout~\cite{srivastava2014dropout} being a ubiquitous example.  Another example is the so-called \emph{reparametrization trick} proposed by \cite{kingma2013auto,rezende2014stochastic} 
for backpropagation through certain stochastic units
(e.g., Gaussian) that allow one to differentiate quantities of the form $\mathbb{E}[f(x)]$ where $x\sim P(x; \theta)$ with respect to distribution parameters $\theta$.
Schulman~\ea~\cite{schulman2015gradient} recently united the REINFORCE and reparameterization tricks to handle backpropagation on general ``stochastic computation graphs’’.

\paragraph{Motion prediction as Inverse Optimal Control (IOC)}
Works of \cite{ziebart2009planning,kitani2012activity,huang2015approximate} 
model pedestrians as rational agents that move in a way that optimizes a reward function dependent on image features, e.g., obstacles in the scene, and form a distribution over optimal future paths to follow. They train a reward function 
such that the resulting Markov Decision Process  best matches  observed human trajectories in expectation. 
Many times though our motions/behaviours are not goal oriented and their framework fails to account for those. In these cases,  learning the possible future outcomes directly from data is important, as noted also in \cite{ziebart2010modelingB}. However, to deal with much longer temporal horizons we do expect that combination of planning and learning to be beneficial. 

\section{Conditional Stochastic Networks }\label{sec:model}

Figure~\ref{fig:Pnets} depicts our three stochastic network models for the three tasks we consider: (1) object trajectory forecasting, (2) body joint trajectory forecasting and (3) video prediction.  Samples  from our models correspond to  plausible future motion  trajectories  of the object,  future body joint trajectories of a person or future frame optical flow fields, respectively, for the three tasks we consider. 


\paragraph{Object-centric prediction}
For dense pixel motion prediction,  visual glimpse correspond to the whole video frame.
For the case of object and body joint  trajectory forecasting however,   visual glimpses are 
centered around the object of interest. 
Such object-centric glimpses  
have been shown to generalize better to novel environments~\cite{fragkiadaki2015learning, DBLP:journals/corr/BattagliaPLRK16}, as they provide translation invariance. Further, the same forecasting model can be shared by all objects in the scene to predict their future trajectories.

\begin{figure}
\centering
\includegraphics[width=0.99\linewidth]{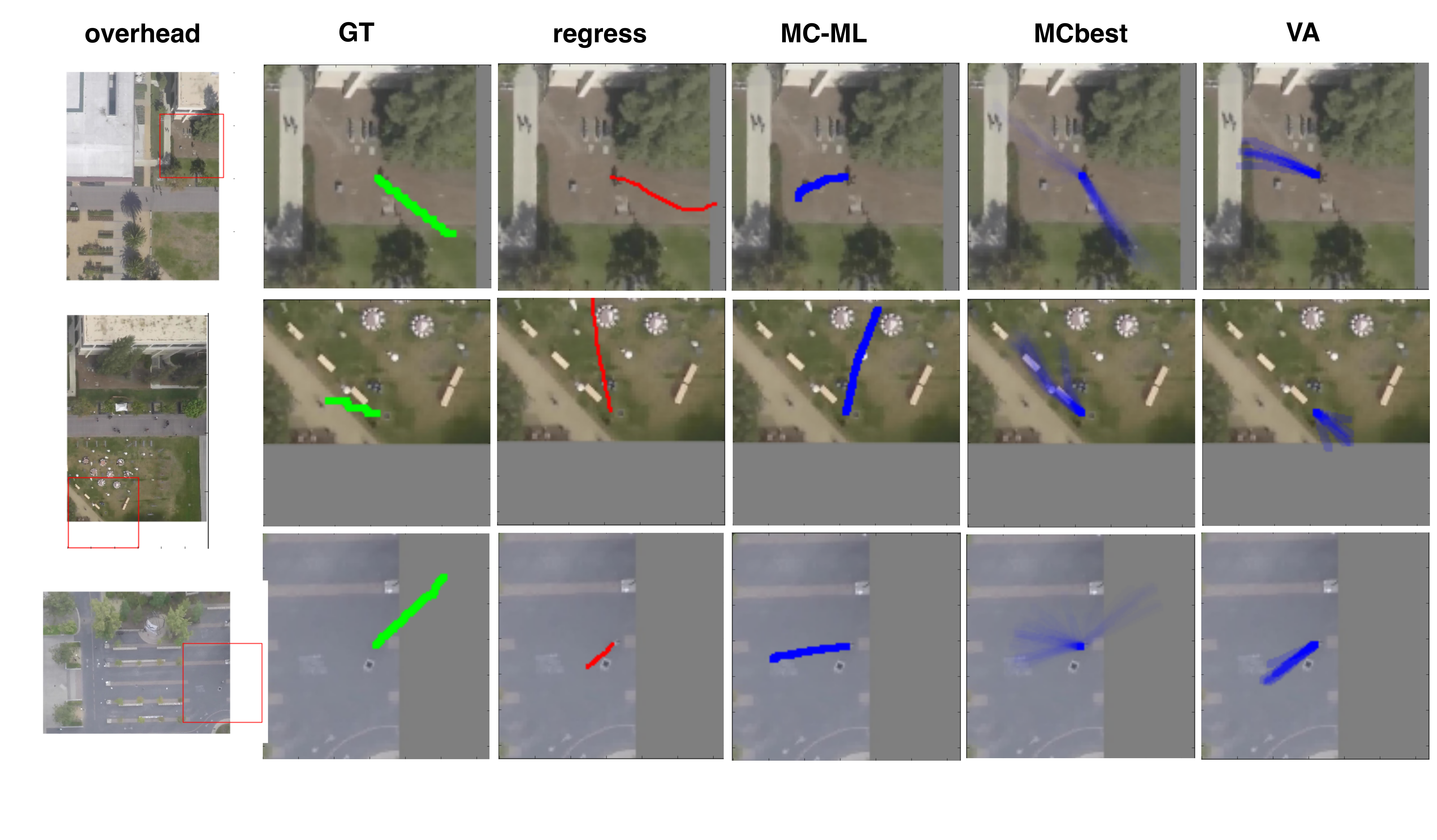} 
\caption{\textbf{Object motion trajectory  prediction} in the drone Stanford dataset of \cite{DBLP:journals/corr/RobicquetASADWS16}.   A single frame is used as input ($\Nf=1$).  This is a case of \textit{extreme uncertainty}:  we do not have any idea regarding the direction of object's motion.  The first column shows the overhead scene and the input visual glimpse and second column shows the groundtruth trajectory. Dark blue denotes high sample density and light blue denotes low sample density.  MCbest predicts more diverse outcomes (e.g., row 3) in comparison to the rest.  
}
\label{fig:alahiqual}
\end{figure}

Our model has three components: an encoder $\encoder$, a stochastic layer $\zz$, and a decoder $\decoder$.  
The encoder is a deep convolutional neural network that takes as input a short sequence of visual glimpses $\xx$ and  computes a  feature hierarchy. The  stochastic layer is a $\ncodes$-dimensional  random variable $\zz$ drawn from a fully factored Gaussian distribution, whose means and standard deviations are parametrized by the top layer activations of the encoder $\encoder(\xx)$ through linear transformations $(\MW,\Mb)$ and $(\SW,\Sb)$:

\begin{align*}
\mu(x)&= \MW \cdot \encoder(\xx; \theta_\encoder) +\Mb, \\
\sigma(x)&= \log(\exp(\SW \cdot \encoder(\xx; \theta_{\encoder}) +\Sb)+1), \\
\zz &\sim \mathcal{N}(\mu, \mbox{diag}(\sigma)),
\end{align*}
For the standard deviations  we use a softplus nonlinearity to ensure positivity.  
The decoder $\decoder$ transforms samples of  $\zz$ (as well as encoder feature maps when skip-layers are present) to produce output $\yout$. While each of our stochastic variables $\zz$ follow a (unimodal) Gaussian distribution, 
the  distribution of $\yout$ can in general be highly multimodal as
shown qualitatively in Figure~\ref{fig:alahiqual} since Gaussian samples go through multiple non-linear layers. 


The advantage of using a normal distribution for stochastic units is that back-propagation through the stochastic layer can be performed 
efficiently via the so-called \emph{reparametrization trick}~\cite{kingma2013auto,rezende2014stochastic}, where $\hat{\zz}$ denotes stochastic samples: 
 \begin{align}
\hat{\zz} &=\mu(x)+ \epsilon' \cdot \sigma(x), \;\epsilon' \sim \mathcal{N} (0,I), \nonumber
\end{align}
%
 Given a visual glimpse sequence $x$, we obtain samples of future outcomes $\yhat $   by sampling our stochastic variables $\zz$:
 \begin{align*}
\hat{\zz} &=\mu(x)+ \epsilon' \cdot \sigma(x), \;\epsilon' \sim \mathcal{N} (0,I), \\
\yhat &= \decoder(\hat{\zz}; \encoder(\xx; \theta_\encoder) ; \theta_\decoder),
\end{align*}
%
We train our model using maximum likelihood. 
We employ the standard assumption that ground-truth future outcomes $\yout$ are normally distributed  with means that depend on the output of our model. Let $\theta$  denote the weights of our stockastic network model, then we have:
\begin{align}
\yout &\sim \mathcal{N}(\decoder(\zz;\encoder(x;\theta_\encoder); \theta_\decoder),\nu  I). 
\end{align}
and maximize the marginal log likelihood $\mathcal{L}$ of the 
observed future outcomes conditioned on the input visual glimpses, as previous works:
\begin{small}
\begin{align}
&\mathcal{L}(\theta) \equiv \frac{1}{\N} \sum_{i=1}^\N \log  P(\yout^{(i)}) \equiv 
\frac{1}{\N} \sum_{i=1}^\N \log \int_{\zz^{(i)}} P(\yout^{(i)}, \zz^{(i)} | \xx^{(i)}) d\zz^{(i)} \nonumber \\
&\equiv   \frac{1}{\N} \sum_{i=1}^\N \log \int_{\zz^{(i)}} P(\yout^{(i)}, | \xx^{(i)},\zz^{(i)}) P(\zz^{(i)}|\xx^{(i)}) d\zz^{(i)} \label{eqn:marginalization2}
\end{align}
\end{small}
The marginalization across latent variables in Eq. \ref{eqn:marginalization2} is intractable. 
We will use the following three approximations: (1) a sample approximation to the marginal log likelihood by sampling from the prior (MCML), (2) variational approximations (VA) and (3) $K$-best-sample loss, as detailed below.


\paragraph{Sampling from the prior (MCML)}
We maximize an approximation to the marginal log likelihood $\mathcal{L}$  using a set of $\nsamples$ samples for each of our $\N$ training examples: 

\begin{small}
\begin{align}
&\mathcal{L}^{MCML}(\theta) \equiv   \frac{1}{\N} \sum_{i=1}^\N \log \sum^\nsamples_{\substack{j=1 \\ \hat{\zz}^{(i,j)} \sim P(\zz|\xx^{(i)})}} P(\yout^{(i)}| \hat{\zz}^{(i,j)}, \xx^{(i)}),\label{eqn:stochastic1} \\
    &=  \frac{1}{\N} \sum_{i=1}^\N \log \sum_{j=1}^\nsamples \exp\left( -\frac{\|\yhat^{(i,j)} - \yout^{(i)} \|^2}{2 \nu}\right) \label{eqn:logsumexp} 
\end{align}
\end{small}
We use $\nsamples=15$ samples for each training example.

\begin{figure}[h!]
\begin{center}
\includegraphics[width=0.45\textwidth]{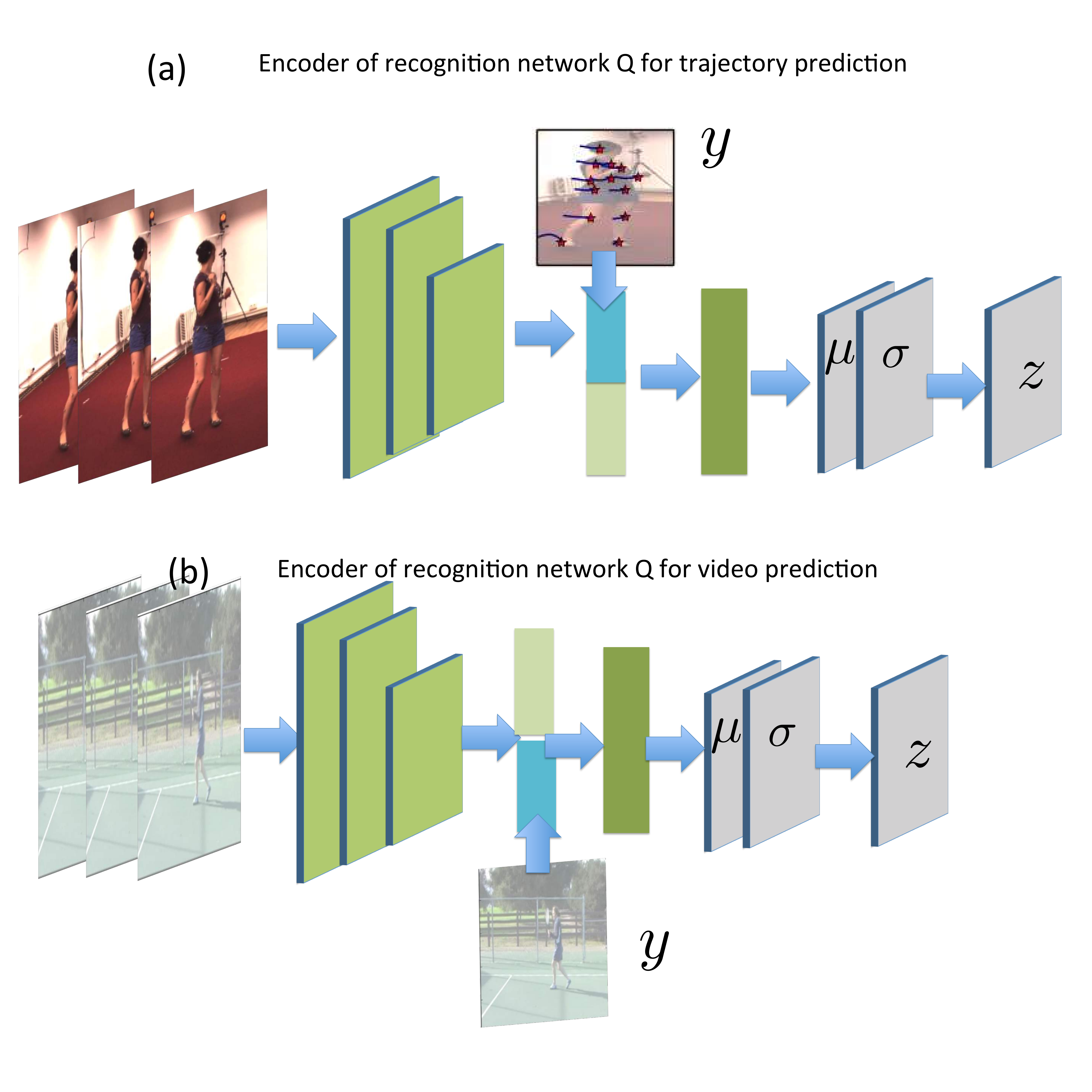}
\end{center}
\caption{
\textbf{Recognition networks $Q$} for the tasks of body joint prediction and video prediction. The output $\yout$ is supplied as input to the model to provide an informative distribution over stochastic $\zz$ codes which our generator network $P$ tries to approximate.
}
\label{fig:Q}
\end{figure}

\paragraph{Variational approximation (VA)}
With a stochastic encoder and multimodal predictions, the likelihood of our model is sensitive to the sampling of the stochastic codes, which means, we would require a \textit{large amount of samples} to provide a good approximation to the marginal likelihood, which is the disadvantage of using MCML.  Another alternative is to do importance sampling in the stochastic codes, by taking into account the likelihood of each one.  
Returning to Equation~\ref{eqn:marginalization2} from the previous section, but doing importance sampling over our stochastic codes, our loss takes the form:
{\small
\begin{align}
& \mathcal{L}(\theta) =
\frac{1}{\N} \sum_{i=1}^\N \log \int_{\zz^{(i)}} P(\yout^{(i)}, | \xx^{(i)},\zz^{(i)}) \nonumber \\
&\qquad \times \frac{P(\zz^{(i)}|\xx^{(i)})}{Q(\zz^{(i)} | \xx^{(i)} \yout^{(i)})}
Q(\zz^{(i)} | \xx^{(i)} \yout^{(i)}) d\zz^{(i)}, \label{eqn:mliw} \\
&\geq 
\mathcal{L}^{VA}(\theta) =
\frac{1}{\N} \sum_{i=1}^\N  \int_{\zz^{(i)}} 
Q(\zz^{(i)} | \xx^{(i)} \yout^{(i)}) d\zz^{(i)} \nonumber \\
& \qquad \times \log
P(\yout^{(i)}, | \xx^{(i)},\zz^{(i)})
\frac{P(\zz^{(i)}|\xx^{(i)})}{Q(\zz^{(i)} | \xx^{(i)} \yout^{(i)})} , \label{eqn:vaelower} \\
&= \frac{1}{\N} \sum_{i=1}^\N 
\mathbb{E}_{\zz\sim Q(\zz^{(i)} | \xx^{(i)} \yout^{(i)})} 
\left[ \log P(\yout^{(i)}, | \xx^{(i)},\zz^{(i)})\right] \nonumber \\
& \qquad - \textrm{D}_{\textrm{KL}} \left( Q(\zz^{(i)} | \xx^{(i)} \yout^{(i)}) \mid \mid P(\zz^{(i)} | \xx^{(i)}) \right), \label{eqn:vaeobj}
\end{align}\\
}
Equation~\ref{eqn:vaelower} is a lower bound of the importance 
weighted maximum likelihood objective (Equation~\ref{eqn:mliw}),
formed by taking the log of an expectation as the expectation of a log
using Jensen's inequality.
This requires creating a new approximation ($Q$) to the true posterior.  Here $Q$ attempts to give us
the best $\zz$ that can create a particular path $\yout$. 
In our final objective in Equation~\ref{eqn:vaeobj} we see that this
takes the form of a variational autoencoder style objective, as first noted in \cite{NIPS2015_5775}; since our $Q$ network further conditions on input $x$ (rather than just output $y$), it is called \textit{conditional} variational autoencoder.    
In the term on the left, we are just forming
maximum likelihood predictions, but using codes from our recognition network $Q$ (we found it sufficient to use only one sample per training example and iteration) instead of our generator, while the term on the right forces the distribution of the stochastic codes of our  recognition  network $Q$  and our generator to be close (in KL divergence terms).  
The architecture of our recognition networks for the tasks we consider are illustrated in Figure \ref{fig:Q}. At test time, of course only the generator model is used.


\paragraph{$K$-best-sample-loss (MCbest)}
We introduce $K$-best-sample-loss $\mathcal{L}^{best-sample}(\theta)$ that takes into account only the best sample within a budget of $\nsamples$ predictions, for each training example: 
\begin{small}
\begin{align}
& \mathcal{L}^{best-sample}(\theta) \equiv   \frac{1}{\N} \sum_{i=1}^\N \max_{\substack{j=1 \cdots \nsamples \\ \hat{\zz}^{(i,j)} \sim P(\zz|\xx^{(i)})}} P(\yout^{(i)}| \hat{\zz}^{(i,j)}, \xx^{(i)}),\label{eqn:stochastic} \\
    &\equiv  \frac{1}{\N} \sum_{i=1}^\N 
    \min_{j=1\cdots \nsamples} \| \yout^{(i)} - \yhat^{(i,j)} \|^2 
    \label{eqn:bestsample}
\end{align}
\end{small}
Intuitively, for each training example during training, our model makes $\nsamples$ guesses (i.e. draws $\nsamples$ samples) about  the future outcome, not all of which have to be ``correct'' --- we only penalize the best guess made by the model for its deviation from ground-truth.  This approach consequently
encourages our model to ``cover its bases’’ by making diversified predictions. 


\begin{figure*}
\centering
 \includegraphics[width=1.0\linewidth]{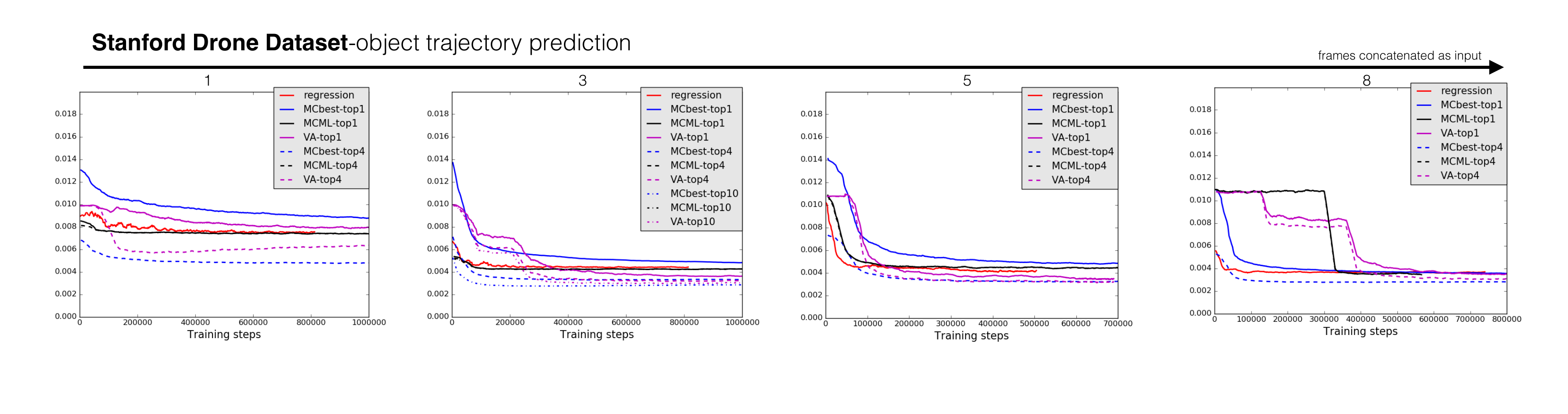} 
\caption{
\textbf{Object trajectory prediction} in the Drone Stanford  dataset~\cite{DBLP:journals/corr/RobicquetASADWS16} under varying number of  input frames $\Nf$.  We show top$k$ error,  for $k=1$ and $k=4$ . \textit{Col 1:} Forecasting future motion trajectory from a single frame.  This is a case of extreme uncertainty and MCbest outperforms the rest on top-4 error. 
When a long visual history is used as input ($\Nf=8$), the benefit of stochastic models  over a  deterministic regression baseline is smaller, as uncertainty in prediction decreases (column 4). 
}
\label{fig:dronecurves}
\end{figure*}

\begin{figure}
\centering
 \includegraphics[width=1.0 \linewidth]{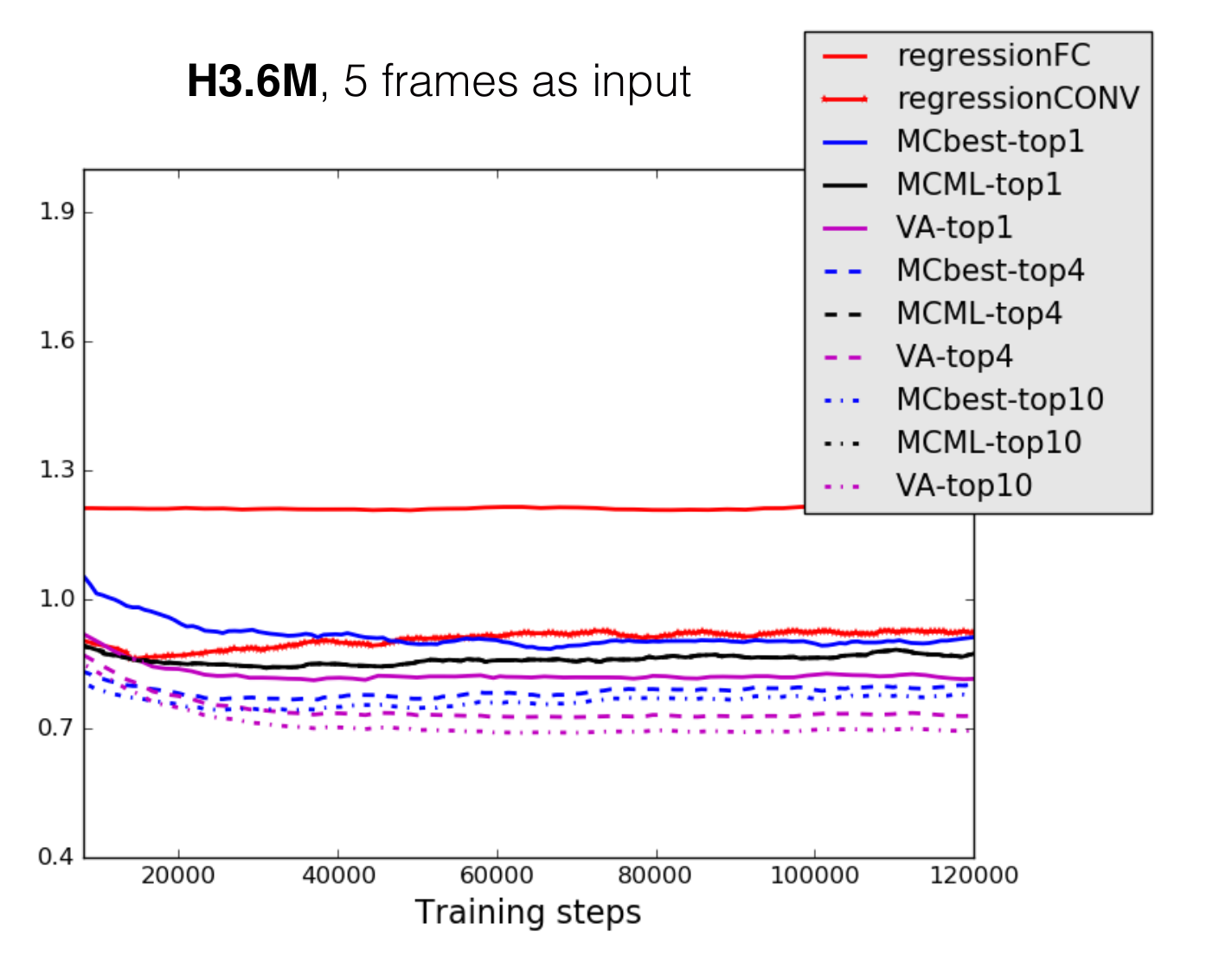} 
\caption{ \textbf{Body joint trajectory forecasting} in the H3.6M test set. All actions categories are considered and five frames are used as input ($\Nf=5$). All models apart from \textit{regressionFC} share the same architecture that use  conv-indexed decoders. The regression baseline \textit{regressionFC} has a fully-connected decoder instead for comparison.  \textbf{Fully connected decoders  under-fit our dataset and failed to minimize the error}. For clarity we show only the fully connected regression model since the stochastic fully connected models exhibit similar behavior.  Conditional variational autoencoders outperform all other training schemes in this experiment.  }
\label{fig:fcconvcurves1}
\end{figure}

\begin{figure}
\centering
 \includegraphics[width=1.0 \linewidth]{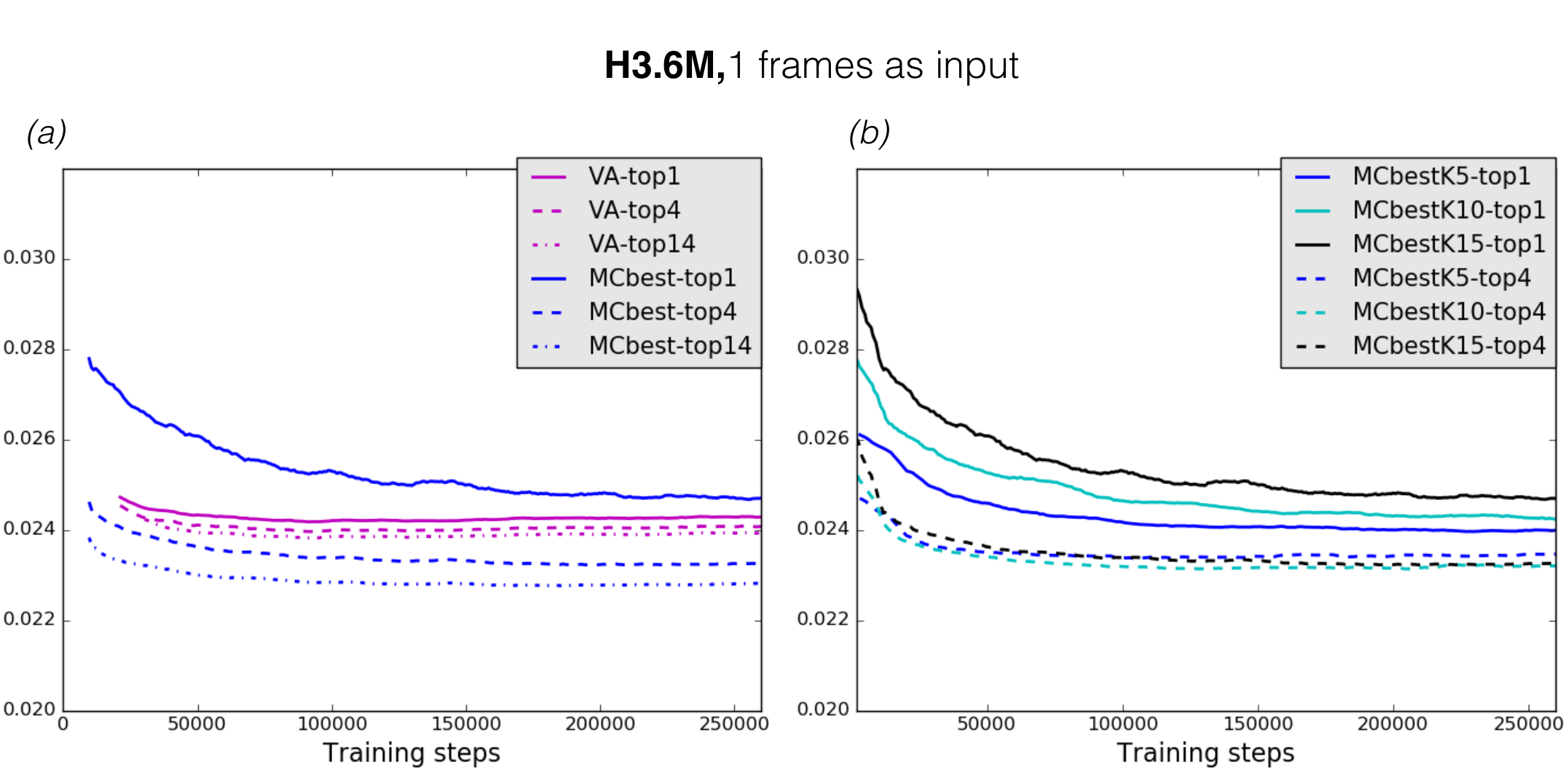} 
\caption{ \textbf{Video prediction} in the H3.6M test set from a single image ($N=1$). \textit{(a)} We compare $15$-best-loss (\textit{MCbest})  against conditional variational approximations (\textit{VA}). $15$-best-loss has higher top1 error than VA but lower top4 and top14 error. \textit{(b)} $K$-best loss under varying $K$: as we increase number of samples $K$, top1 error gets worse and top4 decreases, as expected. $K$ is a hyper-parameters and depends on the amount of anticipated uncertainty.}
\label{fig:fcconvcurves2}
\end{figure}

\begin{figure*}
\centering
 \includegraphics[width=1.0\linewidth]{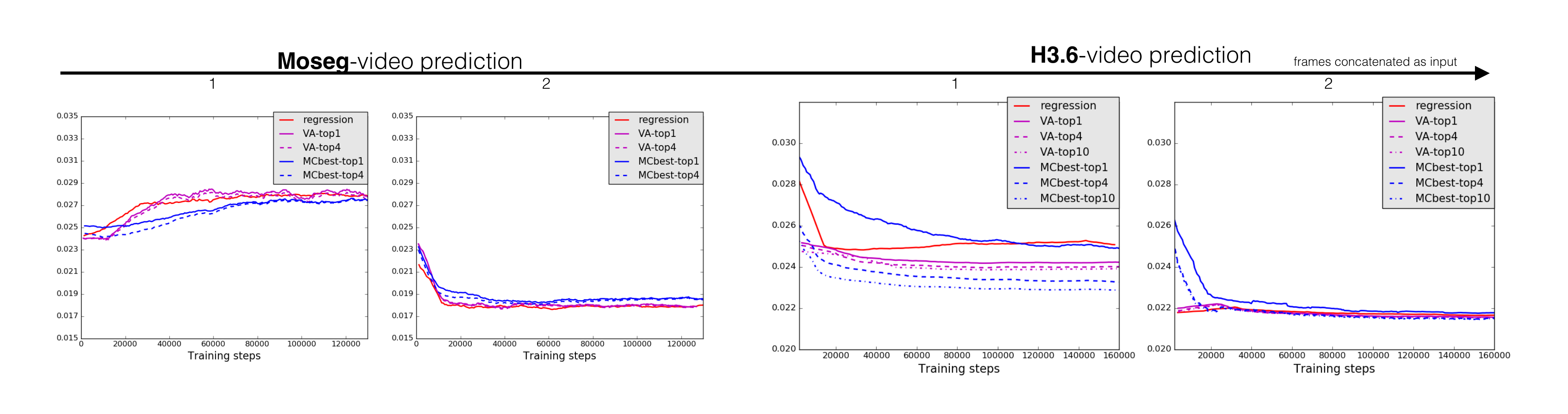} 
\caption{ \textbf{Video prediction} results in MoSeg and H3.6M datasets. 
}
\label{fig:videopredictcurves}
\end{figure*}

\section{Experiments}

We test the performance of our stochastic network models in three tasks: (a) forecasting future motion of human body joints, (b) forecasting video frames, and (c) forecasting future object motion from overhead cameras, and compare against losses and architectures considered in previous works. 
We  compare the following training schemes presented in Section \ref{sec:model}: (a)  a sample approximation to the marginal likelihood (\textit{MCML}), (b) $K$-best-sample loss (\textit{MCbest}), (c) conditional variational autoencoder (\textit{VA}) for the various architectures we consider in each task. We also consider   a regression model (\textit{regression}) that has similar architecture to the stochastic alternatives in each experiment  but that does not have any stochastic units and  minimizes a regression loss. 

\paragraph{Evaluation metric} We  quantify how well our models perform on predicting the distribution of the future outcomes in our datasets, let it be object trajectories, body joint trajectories or future frames, using top$k$ error, $k \in 1 \cdots 15$: top$k$ error measures the lowest of the errors (against ground-truth) among the top $k$ predictions, where predictions are ordered according to model's confidence. For our stochastic networks, we obtain multiple predictions by sampling  $\zz$  (without ensuring diversity) and decoding it to $\hat{yout}$  and ordering the predictions according to the $\zz$ probability  (in decreasing order) while keeping the top $k$ most confident. Top$k$ error is a meaningful measure in case of uncertainty, when there is no one single best solution, and has been employed widely, e.g.,  in the Imagenet Image classification task \cite{deng2009imagenet}, which has potentially much less uncertainty than our forecasting setup. For regression models, we only have one prediction available and thus all top$k$ errors are identical.

\begin{figure*}
\centering
\includegraphics[width=0.99\linewidth]{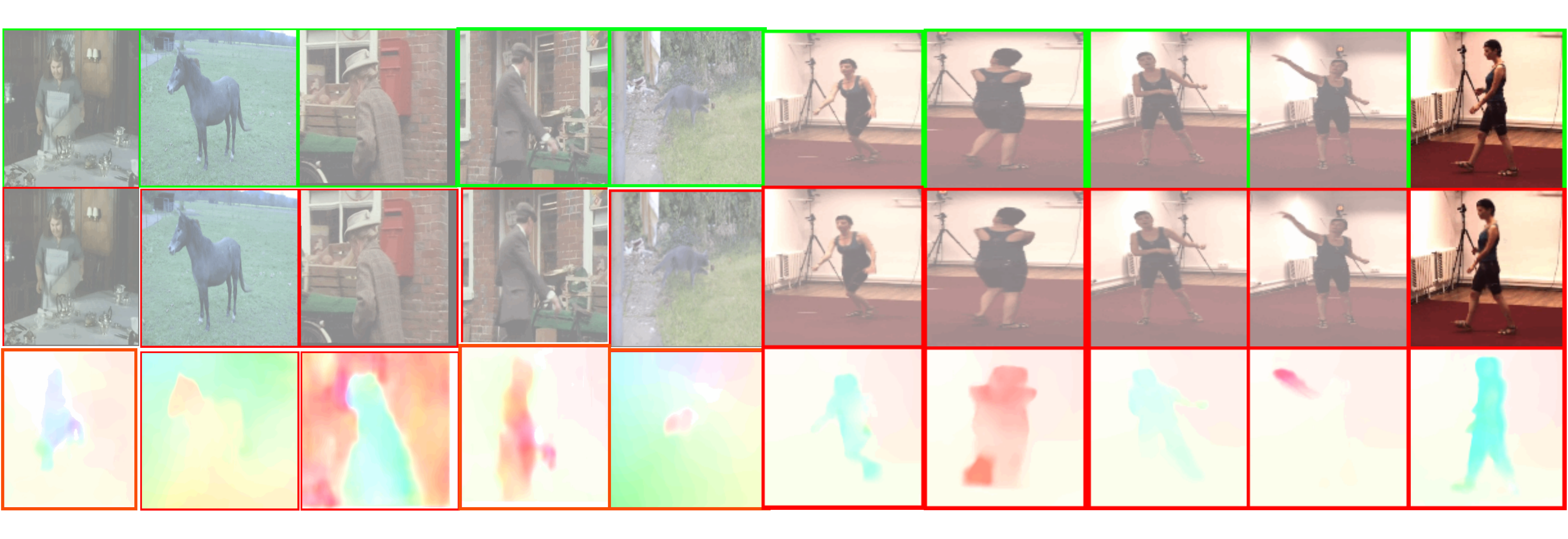} 
\caption{\textbf{Video prediction} in MoSeg (first five columns) and H3.6M (last five columns) using $Nf=2$ frames as input. \textit{Top row:} last frame of visual history. \textit{Bottom Row:} predicted flow field. \textit{Middle Row:} predicted frame by warping. The forecasted flow fields correctly delineate the moving objects. The visual realism of the predicted frames is a benefit of using motion driven differentiable warping  \cite{DBLP:journals/corr/ZhouTSME16}   as opposed to direct pixel  generation through standard deconvolution networks.} 
\label{fig:videopredres}
\end{figure*}

\paragraph{Human body joint trajectory forecasting}
We use the Human3.6M (H3.6M) dataset of Ionescu et al. \cite{h36m_pami}. H3.6M contains videos of actors performing activities and provides annotations of body joint locations at every frame, recorded from a Vicon system. We split the videos according to the human subjects, having subject with id=5 as our test subject. 
We consider videos of all activities. We center our input glimpses around the human actor by using the bounding box provided by the dataset on the latest history frame and crop all history images with the same box (so that only the human moves and the background remains static, as is the case also in the original frame). 
 For training, we pretrained the encoder for body joint detection in the same dataset, and then finetuned the whole model in the forecasting. We consider two types of decoder subnetworks: a four layer fully connected decoder considered in \cite{DBLP:journals/corr/WalkerDGH16} and a conv-indexed decoder, described in Section \ref{sec:model}. 

Our evaluation criterion is L2 distance between predicted and ground-truth instantaneous velocities of all body joints, averaged over the time horizon of the prediction. We use $\Hor=15$.  
We show quantitative results in Figure \ref{fig:fcconvcurves1} (a) for $\Nf=5$ frames in the visual history input to the model. All models with fully connected decoders failed to minimize the error in our experiments and  under-fit the training set. We show in the figure only the regression baseline for clarity (\textit{regressionFC}), as the other curves are identical. Models with  conv-indexed decoders fit the data effectively, and the VA training scheme performed the best.  For extensive qualitative results please see our   \href{https://sites.google.com/site/conditionalstochasticnets/}{online website}.

Conv-indexed decoders assume we know the pixel coordinates $(x,y)$ of the joints whose trajectories we want to predict in the latest frame of the visual history. Such knowledge  is indeed available, without which prediction of velocities (accumulated in time to form predicted $(x,y)$ locations) would not be meaningful.  We note that fully connected decoders effectively worked for the much easier, close to deterministic, problem of  predicting body joint trajectories \textit{only for the Walking activity} which suggests that architectural considerations are important as the forecasting problem becomes harder and more diverse. 


\paragraph{Video prediction}
We use  the Freiburg-Berkeley Motion Segmentation (MoSeg) dataset  of  \cite{springerlink:10.1007/978-3-642-15555-0_21} and  the Human3.6M (H3.6M) dataset of  \cite{h36m_pami}. MoSeg  contains high resolution videos of people, animals, vehicles and other moving objects; the camera  may be moving or static and the frame rate varies from video to video. H3.6M contains videos of a static camera and constant frame rate.  
We applied frame-centric video prediction in MoSeg and object-centric prediction in H3.6M (same as above, but this time we predict the whole frame rather than body joint trajectories). 
 In this way, we can test video prediction both in controlled and ``in-the-wild'' setups. In MoSeg, we follow the official train and test split of the dataset and train our video prediction models from scratch. In H3.6M, we use the same split as for body joint prediction. The input to the network is 128 X 320 for MoSeg and 64X64 for H3.6M. 
 MCML has shown the worst performance so far and since for frame prediction each sample is a whole decoded frame rather than a small set of trajectories we did not consider MCML for this experiment. 
Due to GPU memory limitations, we could only test 5-best-loss for MoSeg while for H3.6M we test 15-best-loss (since the input to the network is smaller).  
We also compare against a regression baseline that predicts a single flow field and warps to generate the future frame, similar to the model of  \cite{DBLP:journals/corr/PatrauceanHC15}.  

Quantitative results showing the L2 loss between next ground-truth frame and predicted frame for the test set of the two datasets are shown in Figures \ref{fig:videopredictcurves}, \ref{fig:fcconvcurves2} and qualitative results are in Figure~\ref{fig:videopredres}. 
All models fail to minimize the prediction error in MoSeg dataset when only one frame is considered due to camera motion: camera motion dictates the appearance of the next frame and is impossible to be predicted from a single frame. This is one reason previous works use  background subtraction to stabilize the video \cite{DBLP:journals/corr/WalkerDGH16,vondrick2016generating}. On the contrary, in H3.6M  the camera is not moving and a single frame  suffices to make plausible predictions of future flow fields. 
Video results and diverse samples for future frames are available at our \href{https://sites.google.com/site/conditionalstochasticnets/}{online website}  as gifs,  since they are often too subtle to see in a static image.


\paragraph{Object future motion trajectory prediction}
We use the Stanford Drone dataset introduced in Robicquet et al. \cite{DBLP:journals/corr/RobicquetASADWS16}, the largest overhead dataset that has been collected so far, with a drone flying above various locations in the Stanford campus. Trajectories of vehicles, pedestrians, bikers, bicyclists, skateboarders etc, as well as their category labels have been annotated. We split trajectories into train and test set randomly.  We use a short sequence of $\Nf$ square visual glimpses around the target of interest (again following the principle of object-centric prediction) as input to our model and predict future motion for the subsequent $\Hor=100$ frames.   We do not use the annotated category labels. We randomly split object trajectories into train and test sets. 
Our evaluation criterion is L2 distance between predicted and ground-truth instantaneous velocities, averaged over the time horizon of the prediction.

We show quantitative results of our models and a regression baseline in Figure \ref{fig:dronecurves}. The deterministic regression baseline is similar to the model used in \cite{fragkiadaki2015learning} for predicting trajectories of Billiard balls in simulated (deterministic)  worlds. Under uncertainty,  regression  has higher top1  error than VA, and  much higher error than top4 errors of VA and $15$-best-loss, for any number of input history frames $\Nf$ (for $\Nf=1$, VA takes much longer to train than regression). 

\textbf{As the length of the history increases, uncertainty decreases} and regression  becomes more competitive. 
Indeed, most of the times a linear object motion model suffices. It is at the \textit{intersections} that a multimodal predictive model is needed, but these are precisely the cases when good predictions are crucial, e.g., to adapt a self-driving car's behavior based on anticipated pedestrian motion.  
Please see our \href{https://sites.google.com/site/conditionalstochasticnets/}{online website} for video results.



\subsection{Implementation details.} \label{sec:details}
Our encoder $\encoder$ is a convolutional network variant of the Inception architecture~\cite{szegedy2015going}. We pre-train our encoders on the ImageNet classification task by replicating the weights of the first convolutional layer as many times as the number of frames in the input glimpse volume. For body joint trajectory forecasting, we pretrain our encoder on body joint classification in H3.6M dataset. 
We use skip layers in the tasks of body joint trajectory forecasting and video forecasting.
The stochastic variables are replicated across the width and height of the top encoder layer in case of fully convolutional decoders. 
We represent output motion trajectories with a sequence of instantaneous velocities $\tr=\{(\Delta{\trx_t}, \Delta{\try_t}), \ t=0,\dots,\Hor\}$, where $\Hor$ is a prediction horizon. 
We conduct our training and evaluation using TensorFlow~\cite{citeulike:13838961}, a publicly available deep learning package. 

\section{Conclusion}
\label{sec:conclusion}

We presented stochastic neural networks for multimodal motion forecasting for object, body joint and frame predictions. We presented architectural innovations as well as novel training schemes and extensive evaluations of those in various benchmarks. Our work aims to help understanding stochastic networks and their applicability for forecasting, and the difficulties (or not) of their training, building upon and extending recent previous works.   
We discussed both object and body joint trajectory prediction that uses human annotations for future outcomes as well as self-supervised video prediction that uses the next frame as ground-truth. The longer the temporal horizon for the prediction, and the shorter the temporal history to condition upon, the larger the uncertainty and the larger the benefit from incorporating stochasticity.  
Our model offers a simple solution to  multimodal prediction of large continuous spaces and we expect it will  be useful in  domains with uncertainty beyond motion prediction. 

\begin{small}
\bibliographystyle{ieee}
\bibliography{egbib}
\end{small}
\end{document}